\documentclass[letterpaper, 10 pt, conference]{ieeeconf}
\IEEEoverridecommandlockouts                              

\usepackage{dblfloatfix}
\usepackage{epsfig}
\usepackage{psfrag,graphicx,xcolor}
\usepackage[normalem]{ulem}
\usepackage{amsmath,amsfonts}
\usepackage[hidelinks]{hyperref}
\usepackage{epsfig,psfrag,graphicx,color}
\usepackage{amsthm}
\usepackage{amssymb}
\usepackage{algpseudocode}
\usepackage{algorithm}
\usepackage{enumerate}
\usepackage{color} 
\usepackage{url} 
\usepackage{cleveref}
\usepackage{booktabs, makecell, tabularx}
\font\bfmath=cmmib10
\mathchardef\Gamma="7100
\mathchardef\Delta="7101
\mathchardef\Theta="7102
\mathchardef\Lambda="7103
\mathchardef\Xi="7104
\mathchardef\Pi="7105
\mathchardef\Sigma="7106
\mathchardef\Upsilon="7107
\mathchardef\Phi="7108
\mathchardef\Psi="7109
\mathchardef\Omega="710A

\mathchardef\alpha="710B
\mathchardef\beta="710C
\mathchardef\gamma="710D
\mathchardef\delta="710E
\mathchardef\epsilon="710F
\mathchardef\zeta="7110
\mathchardef\eta="7111
\mathchardef\theta="7112
\mathchardef\iota="7113
\mathchardef\kappa="7114
\mathchardef\lambda="7115
\mathchardef\mu="7116
\mathchardef\nu="7117
\mathchardef\xi="7118
\mathchardef\pi="7119
\mathchardef\rho="711A
\mathchardef\sigma="711B
\mathchardef\tau="711C
\mathchardef\upsilon="711D
\mathchardef\phi="711E
\mathchardef\chi="711F
\mathchardef\psi="7120
\mathchardef\omega="7121
\mathchardef\epsilon="7122

\mathchardef\varepsilon="7122
\mathchardef\vartheta="7123
\mathchardef\varpi="7124
\mathchardef\varrho="7125
\mathchardef\varsigma="7126
\mathchardef\varphi="7127
\mathchardef\imath="717B
\mathchardef\jmath="717C

\def\bfp{{\mbox{\boldmath $p$}}}

\def\smallbfW{{\raise1.5pt\hbox{\mbox{\boldmath $_W$}}}}

\def\blue{\color{blue}}

\def\mypsfrag#1#2#3#4#5{
        \begin{figure}[htp]
           \begin{center}
              {\leavevmode
                 {\includegraphics[width=#1truecm]{#2.eps}}
              }
           \end{center}
           \vspace{#3}
           \caption{#4}
           \label{#5}
        \end{figure}
}

\def\my4psfrag#1#2#3#4#5#6#7#8{
        \begin{figure}[htp]
        \begin{center}
            \begin{tabular}[h]{c c}
              {\leavevmode{\includegraphics[width=#1truecm]{#2.eps}}}
              &
              {\leavevmode{\includegraphics[width=#1truecm]{#3.eps}}} \\
              {\leavevmode{\includegraphics[width=#1truecm]{#4.eps}}}
              &
              {\leavevmode{\includegraphics[width=#1truecm]{#5.eps}}}
         \end{tabular}
           \vspace{#6}
           \caption{#7}
           \label{#8}
        \end{center}
        \end{figure}
}

\def\mydouble4psfrag#1#2#3#4#5#6#7#8{
        \begin{figure*}[htp]
        \begin{center}
            \begin{tabular}[h]{c c}
              {\leavevmode{\includegraphics[width=#1truecm]{#2.eps}}}
              &
              {\leavevmode{\includegraphics[width=#1truecm]{#3.eps}}} \\
              {\leavevmode{\includegraphics[width=#1truecm]{#4.eps}}}
              &
              {\leavevmode{\includegraphics[width=#1truecm]{#5.eps}}}
         \end{tabular}
           \vspace{#6}
           \caption{#7}
           \label{#8}
        \end{center}
        \end{figure*}
}

\renewcommand{\baselinestretch}{0.96}

\AtBeginDocument{
    \setlength{\belowdisplayskip}{3pt} \setlength{\belowdisplayshortskip}{2pt}
    \setlength{\abovedisplayskip}{2pt} \setlength{\abovedisplayshortskip}{2pt}
}
\makeatletter
\let\origsection\section
\renewcommand\section{\@ifstar{\starsection}{\nostarsection}}

\newcommand\nostarsection[1]
{\sectionprelude\origsection{#1}\sectionpostlude}

\newcommand\starsection[1]
{\sectionprelude\origsection*{#1}\sectionpostlude}

\newcommand\sectionprelude{%
  \vspace{0pt}
}

\newcommand\sectionpostlude{%
  \vspace{0pt}
}

\let\origsubsection\subsection
\renewcommand\subsection{\@ifstar{\starsubsection}{\nostarsubsection}}

\newcommand\nostarsubsection[1]
{\subsectionprelude\origsubsection{#1}\subsectionpostlude}

\newcommand\starsubsection[1]
{\subsectionprelude\origsubsection*{#1}\subsectionpostlude}

\newcommand\subsectionprelude{%
  \vspace{-1.3pt}
}

\newcommand\subsectionpostlude{%
  \vspace{-1.3pt}
}

\let\origsubsubsection\subsubsection
\renewcommand\subsubsection{\@ifstar{\starsubsubsection}{\nostarsubsubsection}}

\newcommand\nostarsubsubsection[1]
{\subsubsectionprelude\origsubsubsection{#1}\subsubsectionpostlude}

\newcommand\starsubsubsection[1]
{\subsubsectionprelude\origsubsubsection*{#1}\subsubsectionpostlude}

\newcommand\subsubsectionprelude{%
  \vspace{1pt}
}

\newcommand\subsubsectionpostlude{%
  \vspace{-1.3pt}
}

\usepackage{etoolbox}
\makeatletter
\patchcmd{\@makecaption}
  {\scshape}
  {}
  {}
  {}
\makeatletter
\patchcmd{\@makecaption}
  {\\}
  {.\ }
  {}
  {}
\makeatother

\makeatother
\usepackage[belowskip=-7pt,aboveskip=10pt,font={small}]{caption}
\usepackage[font={small}]{subcaption}

\def\bfp{{\mbox{\boldmath $p$}}}

\usepackage{amssymb}  

\def\A{\mathcal{A}}
\def\T{\mathcal{T}}
\def\C{\mathcal{C}}
\def\Tu{\widetilde{\T}}

\usepackage{amsmath}

\makeatletter

\makeatletter
\IEEEtriggercmd{\reset@font\normalfont\fontsize{7.5pt}{8.40pt}\selectfont}
\makeatother
\IEEEtriggeratref{1}

\begin{document}

\title{\LARGE \bf  
A Task Allocation Framework for Human \\Multi-Robot Collaborative Settings
}

\author{Martina Lippi${}^1$, Paolo Di Lillo${}^2$ and Alessandro Marino${}^2$
\thanks{${}^1$M. Lippi is with Roma Tre University, Italy,
        {\it martina.lippi@uniroma3.it}}
\thanks{${}^2$P. Di Lillo and A. Marino are with University of Cassino and Southern Lazio,  Italy, {\it \{pa.dilillo,al.marino\}@unicas.it}
        }
\thanks{This work was supported by H2020-ICT project CANOPIES  (Grant Agreement N. 101016906).}        
  }

\maketitle

\begin{abstract}
The requirements of modern production systems together with more advanced robotic technologies have fostered the integration of teams comprising humans and autonomous robots.
However, along with the potential benefits, the question also arises of how to effectively handle these teams considering the different characteristics of the involved agents.
This paper presents a framework for task allocation in a human multi-robot collaborative scenario. The proposed solution combines an optimal offline allocation with an online reallocation strategy which accounts for inaccuracies of the offline plan and/or unforeseen events, human subjective preferences and cost of task switching, so as to increase human satisfaction and team efficiency.
Experiments with two manipulators cooperating with a human operator in a box filling task are presented.
\end{abstract}

\section{Introduction}\label{sec:intro}
Human-Robot Collaboration (HRC) has become a key technology in modern production systems to achieve flexibility and high quality: the reasoning abilities, advanced perception and dexterous manipulation of the former are combined with the endurance, precision and strength of the latter.
This opens up the problem of how to \emph{optimally assign} tasks to the two types of agents considering their inherently different features~\cite{Hashemi_ARC2020}. 
Although optimal task allocation is a well-known problem for multi-robot systems~\cite{Khamis_Springer2015}, there are fundamental differences between  human-robot and multi-robot scenarios:  humans are characterized by several parameters that are generally challenging to  quantify, interdependent, time-varying, affected by external factors and different from person to person. 
This makes evident that 
optimal task allocation in this setting is far from trivial, requiring to continuously monitor human and robot requirements and constraints,  and adapt the solution accordingly.

In this context, human-robot optimal task allocation is addressed in~\cite{Zhang_ICRA2020} where human capabilities are monitored and reallocation is performed whenever their variation overcomes a predefined threshold. 
Human-centered task allocation is also considered  in~\cite{Makrini_2019} for  assembly tasks where human capabilities, workload and ergonomics  aspects are taken into account.
An additional example of HRC in industrial setting is presented  in~\cite{Lamon_RAL2019}  with tasks decomposed in simple actions that are   allocated offline based on complexity,  dexterity and required effort. Different features and complexity of tasks are considered in the allocation among human and robots in~\cite{Malik2019ComplexitybasedTA},  showing that this classification lowers deployment and changeover times. Moreover, the stochastic human nature is tackled in \cite{Cheng_RAL2021} where a hierarchical task model from single-agent demonstrations is automatically  built. 

The performance of the human-robot team is at the center of~\cite{Secchi_RAL2021} where task execution constraints,  variability in the task execution by the human and job quality are considered by a scheduling algorithm. 
Human fatigue, as one of the causes of decreased efficiency  and  health damage, is addressed in~\cite{Zhang_RAL2022}.  To counteract its effects in HRC, a task scheduling model that integrates micro-breaks  to  recovery from fatigue accumulation is proposed. 
Human-robot team performance is also the focus  of~\cite{Giele_ICAART2015}. Here, the concept of  Cognitive Task Load (CTL)  is introduced and an adaptive task allocation mechanism is designed that reallocates tasks based on human performance and robot autonomy, while  considering the switching costs to a completely new task plan. 
%
A fluent, efficient and safe collaboration is  the objective of the work in~\cite{Pulikottil_RCIM2021}. The most likely human goals and precedence constraints are inferred  to select the best robot goal and the respective motion plan. 

  This work advances the state of the art by proposing an {\it optimal} task-allocation framework for teams composed by \emph{any} number of robots and humans. The proposed solution allows to quantitatively consider the different nature of robots and humans  by taking into account several features like workload, task quality, and human supervision role as well as  spatial and temporal application constraints. 
 Moreover, the task switching cost is considered that,   
despite being shown to be an important aspect, has been scarcely investigated in the literature for human-robot task allocation. In detail, it 
is a key feature both for humans and  robots, leading to a potential increase of the overall system performance, since it enables avoiding time losses caused by allocating consecutive tasks to agents that are spatially far 
or that are required, for example, to perform time-consuming tool changes. Furthermore, in relation to human satisfaction, reduced switching times potentially lead to a lower CTL and psychological burden due to switching between  tasks of different nature and requiring different skills \cite{Hoffman_THMS2019}.
 %
This work builds on our previous work~\cite{LIPPI_ROMAN2021} in which a Mixed Integer Linear Programming (MILP) formulation was first defined to optimally allocate tasks among humans and robots.
Here, we extend and test this work  in the direction of a more fluent  optimal allocation by introducing the following points:
\begin{itemize}
\item 
The concept of  cost of switching from one task to another is formally introduced and taken into account to derive the optimal allocation; 
\item The possibility to specify  human preferences regarding tasks that the human wish or not to execute 
is provided and   integrated in the MILP formulation;  
\item An online monitoring and update of the above parameters is foreseen enabling re-planning  whenever a new human preference is expressed or the overall performance of the running plan, also involving other relevant task parameters,  decreases below a given threshold; 
\item A validation campaign is carried out to show the effectiveness of the improved formulation compared to~\cite{LIPPI_ROMAN2021}; 
\item  Real-world experiments involving two manipulators and a human agent are shown to validate the framework. 
\end{itemize}


\section{Human multi-robot collaborative setting}\label{sec:setting}
Let $\mathcal{A}=\mathcal{A}_h\cup \mathcal{A}_r$ be the set of $n_a$ agents defined as \mbox{$\mathcal{A}=\{a_1, a_2, \cdots, a_{n_a}\}$}, with  \mbox{$\mathcal{A}_h=\{a_{h,1}, a_{h,2}, \cdots, a_{h,n_h}\}$}   the set of $n_h$ human agents, and $\mathcal{A}_r=\{a_{r,1}, a_{r,2}, \cdots, a_{r,n_r}\}$ the set of $n_r$ robotic agents, i.e., $n_a=n_h+n_r$. Let  $\mathcal{T}$ be the set of $m$ tasks  $\mathcal{T}=\{\tau_1, \tau_2, \cdots, \tau_m\}$ to be executed. We consider that the set of task $\T$ is partitioned into $p$ clusters, with $p\leq m$, on the basis of common properties. For instance, it is reasonable to assume that  two tasks encoding pick-and-place operations of the same item type belong to the same cluster. 
By denoting the cluster $i$ as $\C_i$, it holds \mbox{$\T = \C_1 \,\cup\, \C_2 \,\cup \cdots \cup\, \C_p$}. We introduce the  function $c:\T\rightarrow \C$, with $\C=\{C_1,\cdots,C_p\}$, that provides the cluster associated with a given task. We refer to tasks belonging to the same cluster as \emph{similar}. 
Note that this partitioning  does not compromise the approach generality, since clusters can be associated to unique tasks if no common features exist. 

In the considered collaborative scenario, we envisage that each task must be performed by an agent $a\in\A$, either robotic or human, and that, if necessary, this can also be \emph{supervised} by a human operator. During supervision, the human monitors the task execution and can promptly intervene to guarantee a correct task completion if any anomalies or malfunctions are detected. 
We consider that each task $i\in\T$ 
is characterized by the following \emph{task} parameters: 
\begin{enumerate}
    \item \label{p:loc} estimated \emph{spatial location} $\bfp_i\in\mathbb{R}^3$ where the task is executed, e.g., centroid of the occupation area when performing the task;
    \item estimated \emph{execution time} $\Delta_{i,j}\in\mathbb{R}^+$ to carry out the task $i$ by  agent~$j$, $\forall j\in\A$. This parameter can be set to an arbitrary high value $M$ in case the agent is not able to perform the task. For instance, a robot may be inadequate to realize a task requiring strong dexterity skills like the manipulation of highly deformable objects;
    \item \emph{switching cost} $\Delta^t_{i,k,j}\in\mathbb{R}^+$ representing the time needed by each agent~$j$ to transition  from task $i$ to a subsequent task $k$, $\forall  k\in\T,\, j\in\A$. This cost can include, for example, the time to change tools for the next task $k$ or to travel to the next location $\bfp_k$;  
    \item estimated \emph{execution quality} $q_{i,j}\in[0,1]$  representing a measure of the accuracy achieved by the agent~$j$ in completing the task, $\forall j\in\A$. For example, in a pick-and-place task, this index can quantify the positioning accuracy reachable  by the agent when executing it; 
    \item estimated \emph{supervision quality} $q^s_{i,j}\in[0,1]$ assessing a measure of the accuracy achieved if the human agent~$j$ supervises the execution of the task and possibly intervenes if needed, $\forall j\in\A_h$;
    \item \label{p:w} estimated \emph{workload} $w_{i,j}\in[0,1]$ for each agent~$j$ to carry out the task, $\forall j\in\A$. This variable can represent, for instance, the overall control effort to accomplish the task in case of robotic agent or the cognitive and/or physical effort, e.g., \cite{rouse1993modeling,bommer2018theoretical}, in case of human agent. 
\end{enumerate}
Since similar tasks belong to the same cluster, we assume that same execution and supervision  quality indices are associated with tasks belonging to the same cluster, i.e.,   if \mbox{$c(\tau_i)=c(\tau_k)$}, it holds  $q_{i,j}=q_{k,j}$,  $\forall j\in\A$, and $q_{i,j}^s=q_{k,j}^s$, $\forall j\in\A_h$, $\forall i,k\in\T$.
 Moreover, we assume that  quality indices are additive, implying that the overall quality of a task is given by the sum of execution and supervision quality.

The possibility for human operators to express \emph{preferences} in terms of tasks that they wish to execute or not is also included. Each preference is defined as a tuple $(a,\tau,V)$, where $a\in\A_h$ is the human operator specifying the preference on the task  $\tau\in\T$ and $V\in\{0,1\}$ is the value of the preference, that is $0$ if the human does not want to perform the task and $1$ otherwise. The set collecting possible human preference tuples is denoted by~$\mathcal{H}_p$. 
As realistic in collaborative environments, we consider that all the above task parameters as well as human preferences can generally \emph{vary over time} 
(see Sec.~\ref{sec:reall} for details). 

We additionally foresee that the following constraints can be defined: \emph{i)} \emph{precedence} constraints,  representing preconditions for  tasks to be executed, \emph{ii)} \emph{spatial} constraints, representing need for non-simultaneity of tasks as their spatial locations are too close together, and \emph{iii)} \emph{quality} constraints, representing the minimum accuracy expected for each task. To this aim, 
 the binary variables \mbox{$P_{i,k}, D_{i,k}\in\{0,1\}$}, $\forall i,k\in\T$, are defined, where $P_{i,k}$ is equal to $1$ if task~$i$ has to be completed for task~$k$ to start, and is $0$  if there are no constraints in terms of sequentiality  from  task~$i$ to~$j$, and  $D_{i,k}$ is  $1$ if $\|\bfp_i-\bfp_k\|<\epsilon$, $\forall i,k\in\T$, with $\epsilon$ a positive threshold, and is $0$ otherwise. Any other criterion accounting for the  volume   occupation during task execution  can be applied  to define $D_{i,k}$. 
The positive threshold $\underline{q}\in[0,1]$ is also introduced, denoting the minimum required quality.

 Finally, to  handle switching costs, we define the set $\T_a$ comprising $n_a$ \emph{auxiliary} tasks which do not correspond to real tasks of the system, but allow to account for the starting switching cost for each agent~$j$. 
Let $f:\A\rightarrow\T$ be the function that, for each agent~$j$, provides the respective auxiliary task $i$. We specify that these tasks must be executed before any real task, i.e., it holds $P_{i,k}=1, \forall i\in\T_a, k\in\T$, and the respective switching cost $\Delta_{f(a_j),k,j}$ represents the time to transition from the initial configuration of  agent~$j$ to each  task $k\in\T$, $\forall j\in\A$, while zero execution time is set, i.e.,  $\Delta_{f(a_j),j}= 0$,  $\forall j\in\A$. 
The other task parameters are not relevant for the auxiliary tasks. We denote the set given by the union of $\T$ and $\T_a$ as ${\Tu}$, i.e.,  ${\Tu}=\T\cup \T_a$. 

 Based on the above, we are now ready to formulate the main problem addressed in this work. %

 
 \section{Problem Formulation}
 Consider a human multi-robot collaborative setting, with  agents $\A$ and  assigned tasks $\T$, as described in the previous section.
 Let $\underline{t}_i\in\mathbb{R}^+$ and $\overline{t}_i\in\mathbb{R}^+$ be the starting and final times of task~$i$, $\forall i\in\T$. 
Let $X_{i,j}\in\{0,1\}$ be the binary \emph{allocation} variable, $\forall i\in\Tu,j\in\A$, which is $1$ if agent~$j$ is required to perform task~$i$, and is $0$ otherwise, and let $S_{i,j}\in\{0,1\}$, $\forall i\in\T,j\in\A_h$ be the binary \emph{supervision} variable, which is $1$ if human~$j$ is required to supervise the execution of task~$i$, and is $0$ otherwise. 
Our goals are twofold: \emph{i)} to define an optimal assignment of  starting and final times as well as  allocation and supervision variables, fulfilling any system constraints and human preferences while minimizing a performance metric depending on the overall quality, workload and time, 
and \emph{ii)} to guarantee a correct online execution of the tasks, taking into account the variability of the task parameters and human preferences.  
 
 \subsection{Method Overview}\label{sec:overview}
 
 To solve the above problem, we propose the architecture depicted in 
Figure~\ref{fig:overview}. 
\begin{psfrags}
  \def\scal{0.7}  
  \def\scalsmall{0.5}  
  \psfrag{update}[cc][][\scal]{\hspace{0.3cm}\shortstack[c]{5. Param.  \\ Update}}
  \psfrag{inb}[cc][][\scalsmall]{}
  \psfrag{ina}[cc][][\scalsmall]{\shortstack[c]{$\A,\T$}\vspace{2cm}}
  \psfrag{constr}[cc][][\scal]{\hspace{0.3cm}\shortstack[c]{1. Constr.  \\ Definition}}
  \psfrag{plan}[cc][][\scal]{\hspace{0.1cm}\shortstack[c]{3. Traj. \\ Computation}}
  \psfrag{MILP}[cc][][\scal]{\hspace{0.1cm}\shortstack[c]{2. Optimal \\  Solution }}
  \psfrag{check}[cc][][\scal]{\hspace{0.5cm}\shortstack[c]{6. Realloc.  \\ Check }}
  \psfrag{exec}[cc][][\scal]{\shortstack[c]{\hspace{0.2cm}4. Execution}}
\mypsfrag{7.5}{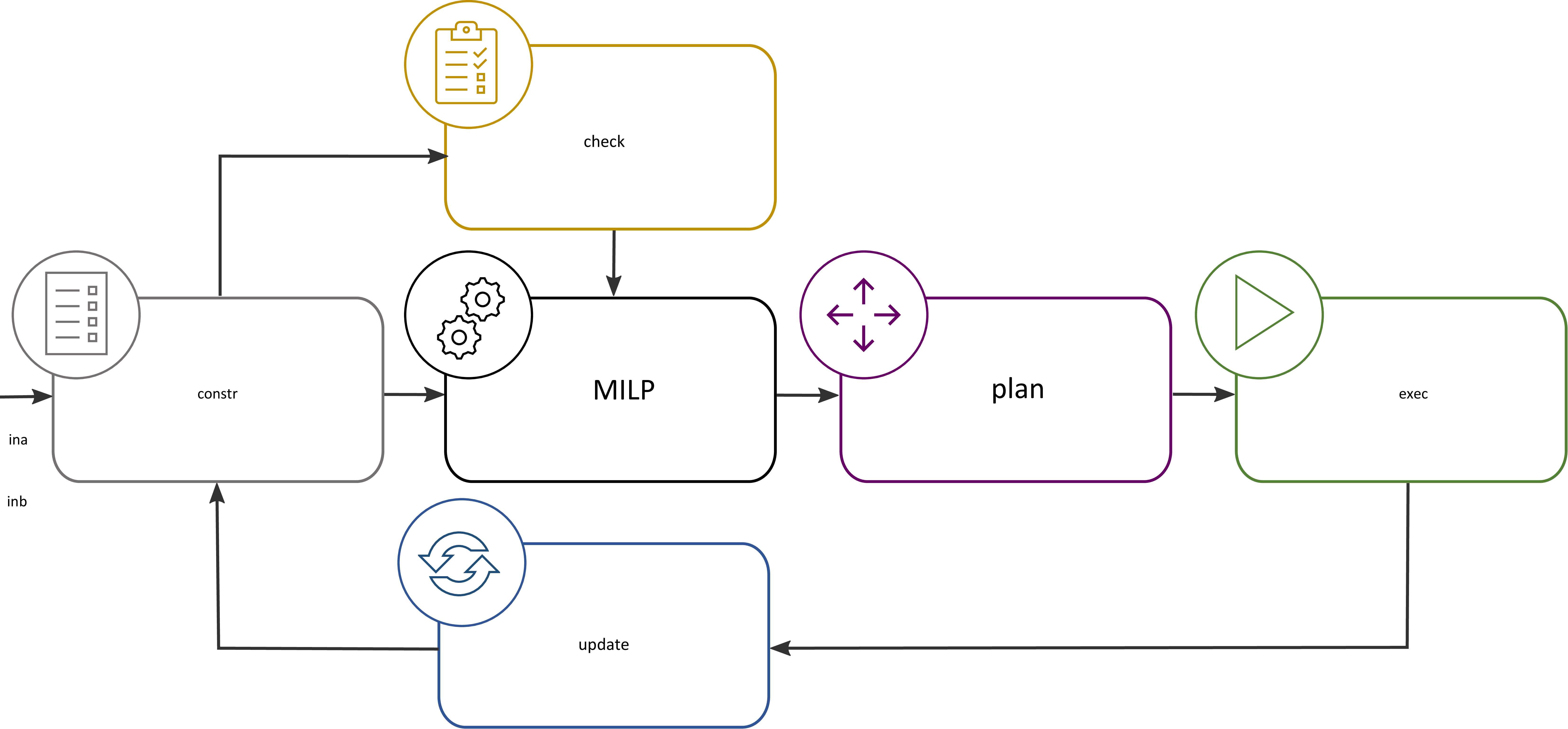}{-12pt}{Overview of the proposed approach.}{fig:overview}
\end{psfrags}
Given the  tasks $\T$ to carry out and the  agents $\A$, we first define a set of constraints (block 1) that must be fulfilled for realizing the tasks (details in Sec.~\ref{sec:milp}). These constraints are used to  formulate an optimal human multi-robot task allocation problem resorting to MILP theory (block 2). By solving this problem, the starting and final times, $\underline{t}_i,\overline{t}_i$, and the allocation, $X_{i,j}$ $\forall i\in\Tu,j\in\A$,  and supervision variables, $S_{i,j}$, $\forall i\in\T,j\in\A_h$, are determined. At this point, the robot trajectories  to accomplish the assigned tasks in the assigned time slots are computed (block 3). Concurrently, humans are informed of the tasks they must carry out or supervise. Then, the task execution  can start (block 4). 

At run-time, as explained in Sec.~\ref{sec:reall}, the task parameters and human preferences are monitored (block 5). Any changes in these are recorded and propagated to  tasks in the same cluster as well as to the set of constraints. 
However, the updated parameters could result in the decision variable assignment no longer meeting the constraints. 
For this purpose, a reallocation module (block 6) is responsible for their checking and for controlling possible impairments in solution optimality. If needed, a new  optimal solution is computed and the robotic and human plans are updated accordingly. 

\section{Optimal Solution}\label{sec:milp}
Let $q$, $w$ and $\overline{t}$ be the overall normalized quality, workload and  time to execute all tasks, respectively, defined~as
\begin{equation}\label{eq:q_w_t}
    \begin{aligned}
    q & = \frac{1}{m}\sum_{i\in \mathcal{T}}\left(\sum_{j\in \mathcal{A}} q_{i,j} X_{i,j}+\sum_{j\in \mathcal{A}_h}q^s_{i,j} S_{i,j}\right) \\
    w & = \frac{1}{m}\sum_{i\in \mathcal{T}}\left(\sum_{j\in \mathcal{A}} w_{i,j}X_{i,j}+\sum_{j\in \mathcal{A}_h}w^s_{i,j}S_{i,j}\right) \\
    \overline{t} & = \max_{i\in\mathcal{T}} \frac{\overline{t}_i}{\Gamma_M}
    \end{aligned}
\end{equation}
with $\Gamma_M$ a makespan upperbound  that can be computed given the estimated execution and switching times. 
We aim to assign the decision variables, i.e., starting and final times and allocation and supervision variables, in such a way that $q$, $w$ and $\overline{t}$ are optimized while fulfilling the system constraints. 
Formally, we propose the following MILP formulation: 
\begin{subequations}  \label{eq:prob}
 \begin{align} 
 \label{eq:obj}
&\min_{X_{i,j}, S_{i,j}, \underline{t}_i, \overline{t}_i}  \quad\overline{t}+w-q\\
& \quad\quad \text{s.t.}\nonumber \\
&\sum_{j\in A}X_{i,j}=1, && \forall i\in\mathcal{T} &  \label{eq:task-ass}\\
& X_{f(a_j),j} = 1, 
 && \forall  j\in\mathcal{A}
 &\label{eq:fake-tasks}\\
  &  X_{i,j} = V, && \forall \,(j, i, V)\in\mathcal{H}_p
 &\label{eq:preferences}\\
&\sum_{j\in \mathcal{A}}q_{i,j} X_{i,j}+\sum_{j\in \mathcal{A}_h}q^s_{i,j} S_{i,j} \geq \underline{q}, && \forall i\in\mathcal{T} &   \label{eq:qual}\\
&  \overline{t}_{i} -\underline{t}_{i} \geq  X_{i,j}{\Delta}_{i,j}   +\sum_{k\in {\Tu}} F_{k,i,j} \Delta^t_{k,i,j},  && \forall i\in \mathcal{T}, j\in\mathcal{A}& \label{eq:duration}\\ 
&\underline{t}_{k} -P_{i,k}\overline{t}_{i} \geq 0, && \forall i,k\in \Tu & \label{eq:precedence}\\ 
&S_{i,j}+ X_{i,j} \leq 1, && \forall i\in\mathcal{T},j\in \mathcal{A}_h  &  \label{eq:supervision}\\
&
\begin{aligned}
&{\underline{t}_k-\overline{t}_i} \ge -M(2-X_{i,j}-X_{k,j} \\
&\quad - S_{i,j} - S_{k,j} ) -M(1-U_{i,k,j})
\end{aligned}
 && \forall i,k\in\mathcal{T}, j\in\mathcal{A} &\label{eq:agent-all}\\
 &
 \begin{aligned}
 &{\underline{t}_i-\overline{t}_k} \ge -M(2-X_{i,j}-X_{k,j} \\
 &\quad- S_{i,j} - S_{k,j} ) -M\,U_{i,k,j}
\end{aligned}
 && \forall  i,k\in\mathcal{T},  j\in\mathcal{A} &\label{eq:agent-all-2}\\
&
\begin{aligned}
&{\underline{t}_k-\overline{t}_i} \ge -M(1-D_{i,k}) \\
& \quad -M(1-Z_{i,k}), 
\end{aligned}
&& \forall i,k\in\mathcal{T} &\label{eq:spatial-all}\\
&
\begin{aligned}
&{\underline{t}_i-\overline{t}_k} \ge -M(1-D_{i,k})  -M\,Z_{i,k},
\end{aligned}
&& \forall i,k\in\mathcal{T} &\label{eq:spatial-all-2}\\
& \overline{R}_{i,k,j} + \underline{R}_{i,k,j} + {R}_{i,k,j} = 1, && \forall i,k\in\Tu,  j\in\mathcal{A} &\label{eq:Rsum}\\
&
 \begin{aligned}
 &\underline{t}_{i}-\overline{t}_{k} - M \overline{R}_{i,k,j} + \underline{R}_{i,k,j} \\ &\quad\leq M(2-X_{i,j}-X_{k,j}),  
\end{aligned}
 && \forall i,k\in\Tu,  j\in\mathcal{A} &\label{eq:Rineq:a}\\
 &
 \begin{aligned}
 & \underline{t}_{i}-\overline{t}_{k} +  \overline{R}_{i,k,j} - M \underline{R}_{i,k,j} \\& \quad\geq -M(2-X_{i,j}-X_{k,j}),
\end{aligned}
 && \forall i,k\in\Tu,  j\in\mathcal{A} &\label{eq:Rineq:b}\\
 &
 \begin{aligned}
 & R_{i,k,j}\leq X_{i,j}, \quad R_{i,k,j}\leq X_{k,j},\\
 & \overline{R}_{i,k,j}\leq X_{i,j}, \quad \overline{R}_{i,k,j}\leq X_{k,j},
\end{aligned}
 && \forall i,k\in\Tu,  j\in\mathcal{A} &\label{eq:Rlimits}\\
 &  \overline{F}_{i,k,j} + \underline{F}_{i,k,j} + {F}_{i,k,j} = 1,  && \forall i,k\in\Tu,  j\in\mathcal{A}
 &\label{eq:Fsum}\\
&
 \begin{aligned}
 & {O}_{k,j}-{O}_{i,j} - M \overline{F}_{i,k,j} + \underline{F}_{i,k,j} \\
 &\quad \leq1 + M(2-X_{i,j}-X_{k,j}),  
\end{aligned}
 && \forall i,k\in\Tu,  j\in\mathcal{A}
 &\label{eq:succ-tasks:a}\\
 &
 \begin{aligned}
 & {O}_{k,j}-{O}_{i,j} +  \overline{F}_{i,k,j} - M \underline{F}_{i,k,j}
 \\ &  \quad \geq 1-M(2-X_{i,j}-X_{k,j}), 
\end{aligned}
 && \forall i,k\in\Tu,  j\in\mathcal{A}
 &\label{eq:succ-tasks:b}\\
  & F_{i,k,j}\leq X_{i,j},\quad  F_{i,k,j}\leq X_{k,j}, && \forall i,k\in\Tu,  j\in\mathcal{A}.
 &\label{eq:Flimits}
\end{align}
\end{subequations}

As mentioned above, the optimization cost in~\eqref{eq:obj} encourages to minimize the overall execution time $\overline{t}$ and  workload $w$, both for executing and supervising tasks, as well as to maximize the overall execution and supervision quality $q$. 

Concerning the constraints, the equality~\eqref{eq:task-ass} states that each task  has to be assigned to an agent.
Similarly, the constraint in~\eqref{eq:fake-tasks} ensures that each auxiliary task in $\T_a$ is assigned to the respective agent, while  human preferences in $\mathcal{H}_p$ are taken into account in \eqref{eq:preferences}.
A minimum task quality $\underline{q}$ for each task $i\in\T$ is required by means of~\eqref{eq:qual} which considers the execution quality index $q_{i,j}$  as well as any  supervision quality $q^s_{i,j}$ provided by humans. 
A minimum task duration according to the allocated agent is established by~\eqref{eq:duration}, where \mbox{$F_{k,i,j}\in\{0,1\}$} is an auxiliary binary decision variable which, as explained in the following, is set to $1$ if tasks~$k$ and~$i$ are both associated with agent~$j$ and task $i$ is consecutive to task $k$, and is equal to $0$ otherwise. In detail, inequality~\eqref{eq:duration} ensures that if  task~$i$ is assigned to  agent~$j$, i.e., it holds $X_{i,j}=1$, then  the assigned task time  $\overline{t}_i-\underline{t}_i$ must be at least  equal to \emph{i)}  the respective estimated execution time $\Delta_{i,j}$  plus \emph{ii)} any switching cost $\Delta^t_{k,i,j}$ for agent~$j$ to transition from a previous task $k$. 

Sequentiality constraints are defined in~\eqref{eq:precedence} which simply imposes that if a task~$i$ has to precede a task~$k$, i.e., $P_{i,k}=1$, then the final time of the former, $\overline{t}_i$, is lower than or equal to the start time of the latter, $\underline{t}_k$. Inequality~\eqref{eq:supervision} ensures that a human \mbox{$j\in\A_h$} cannot simultaneously  execute and supervise a task  $i\in\T$, i.e., either it holds $X_{i,j}=1$ or $S_{i,j}=1$. Moreover, inequalities~\eqref{eq:agent-all}-\eqref{eq:agent-all-2}, with $U_{i,k,j}$ an auxiliary  binary decision variable  $\forall j\in\mathcal{A},$ $i,k\in\mathcal{T}$,  guarantee that each agent can only execute or supervise one task at a time. 
%
Spatial constraints are expressed by~\eqref{eq:spatial-all}-\eqref{eq:spatial-all-2} so that tasks leading agents to work too close together are not  executed simultaneously. Let $Z_{i,k}$ be  an auxiliary binary decision  variable $\forall i,k\in\mathcal{T}$. If a spatial constraint exists between tasks $i$ and $k$, i.e., $D_{i,k}=1$, then  equations \eqref{eq:spatial-all}-\eqref{eq:spatial-all-2} imply that either task $k$ starts after task $i$ is completed, i.e.,  $\underline{t}_k\geq\overline{t}_i$ (if $Z_{i,k}=1$), or the opposite holds true, i.e.,  $\underline{t}_i\geq\overline{t}_k$  (if $Z_{i,k}=0$). No  constraints are imposed if $D_{i,k}=0$. 

Equations~\eqref{eq:Rsum}-\eqref{eq:Flimits} represent the core constraints  to properly set the switching cost auxiliary variable $F_{k,i,j}$ used in~\eqref{eq:duration}. More specifically, in order to
account for switching costs, we need to retrieve the information about whether two tasks, assigned to the same agent, are \emph{consecutive} or not. To this aim, we introduce the auxiliary binary decision variables $R_{i,k,j},\overline{R}_{i,k,j},\underline{R}_{i,k,j}$ and $\overline{F}_{i,k,j},\underline{F}_{i,k,j}$, $\forall i,k\in\Tu,j\in\A$. 

Starting from equations~\eqref{eq:Rsum}-\eqref{eq:Rlimits}, they lead 
$R_{i,k,j},\overline{R}_{i,k,j},\underline{R}_{i,k,j}$ to be representative  of the mutual relationship between the start and final times of pairs of tasks $i$ and $k$ allocated to agent~$j$. By virtue of~\eqref{eq:Rsum},  these variables are mutually exclusive, i.e., only one at a time can be equal to $1$.
Then, according to~\eqref{eq:Rineq:a}-\eqref{eq:Rlimits}, when tasks $i$ and $k$ are allocated  to agent~$j$,  it holds \emph{i)} $R_{i,k,j}=1$, if the start time of task $i$ coincides with the end time of task $k$ ($\underline{t}_{i}=\overline{t}_{k}$), \emph{ii)} $\overline{R}_{i,k,j}=1$, if task $i$ follows task $k$ ($\underline{t}_{i}>\overline{t}_{k}$), and  \emph{iii)} $\underline{R}_{i,k,j}=1$, if task $i$ precedes task $k$ ($\underline{t}_{i,j}<\overline{t}_{k,j}$). No relevant constraints are set if the tasks $i$ and $k$ are not assigned to agent~$j$. Note that the variables $R_{i,k,j},\overline{R}_{i,k,j},\underline{R}_{i,k,j}$ only contain information of whether a task precedes or follows a second task,  but do not capture if they are consecutive. 
This implies that, for example,  all the tasks $k$ of agent~$j$ that start after the completion of a task $i$ will exhibit $\underline{R}_{i,k,j}=1$. 
In view of the above variables, we  define the following integer variable 
\begin{equation}\label{eq:order-task}
    O_{i,j} = \sum\displaystyle_{k\in\Tu}({R}_{i,k,j} + \overline{R}_{i,k,j}), \quad  \forall i\in \Tu, j\in \A 
\end{equation}
which, in case task $i$ is executed by agent~$j$,  provides  its order in the sequence of tasks executed by the same agent. For example, the relation $O_{i,j}=2$ means that  task $i$ is the third task executed by agent~$j$, i.e., two  tasks precede task $i$ for agent~$j$. Based on \eqref{eq:order-task}, we can  determine if two tasks $i$ and $k$ are consecutive  for agent~$j$, with $k$ following~$i$, by verifying if the condition  $O_{k,j}-O_{i,j}=1$ is met or not.  
This is exploited in~\eqref{eq:Fsum}-\eqref{eq:Flimits} to properly set $F_{i,k,j}$. 
In detail, the equality in~\eqref{eq:Fsum} specifies the mutual exclusivity of variables ${F}_{i,k,j},\overline{F}_{i,k,j},\underline{F}_{i,k,j}$. Then, the inequalities in~\eqref{eq:succ-tasks:a}-\eqref{eq:Flimits} implement an \emph{if-then-else} condition such that, if tasks $i$ and $k$ are allocated to agent~$j$ and are consecutive,  with task $k$ following  $i$, then it holds $F_{i,k,j}=1$.

In summary, the proposed optimal formulation can be easily adapted to various human multi-robot  collaborative settings involving different tasks and/or number of agents. 
 Concerning the computational complexity,  it is well-known that  MILP problems are NP-hard. 
 This issue is behind the scope of this work; however,  ad-hoc heuristics can be included to mitigate this issue as, for example, in~\cite{fischetti2010heuristics}.

\section{Online Reallocation}\label{sec:reall}


\subsection{Monitoring and update}
When a task is completed, the monitoring and update phase is activated. 
Let us consider a task~$i$, executed after task $k$ by agent~$j$, and completed by the same agent. Let us also consider that the task may have been possibly supervised by the human~$l$. We refer to the parameters' values prior updating as \emph{nominal} values and assume that all the task parameters~\ref{p:loc})-\ref{p:w}) in Sec.~\ref{sec:setting} can be monitored and measured.  The parameters are updated as follows. 

The spatial location $\bfp_i$, execution $\Delta_{i,j}$ and switching $\Delta_{k,i,j}$ times, and workloads (for executing $w_{i,j}$ and possibly supervising $w_{i,l}^s$) are updated with the current measure in regard to the involved agents. The same parameters, except the spatial location, are also updated for the other tasks in the cluster~$c(\tau_i)$ according to the same proportion. This implies that if, for example,  a certain variation of the execution time is recorded compared to the nominal value, then the same variation is propagated to execution times $\Delta_{q,j}$ of similar tasks $q$ in the cluster $c(\tau_i)$. Updating the spatial location $\bfp_i$ also leads to updating the binary variable $D_{i,q}$, $\forall q\in\T$. 

As far as the quality parameters are concerned, we distinguish two possible cases. In case there is no supervision or the supervising human does not need to intervene, the quality $q_{i,j}$ is updated with the measured one as well as the same update is applied to the similar tasks in  $c(\tau_i)$. In case the task is supervised and the supervising human $l$ has to intervene, the supervision quality $q^s_{i,l}$ is set equal to the measured one and no update is made on  any execution quality. 

Moreover, we consider that the humans can express preferences online for tasks to execute or not execute. Therefore, whenever a new preference is provided or a previous one is modified, the set of preferences $\mathcal{H}_p$ is updated accordingly. 

\subsection{Reallocation  strategy}\label{sec:reall-b}
The strategy for determining whether to reallocate future tasks or not is the following. 
First, we check that the current solution in terms of task starting and final times as well as supervision and allocation variables is still feasible, i.e., no constraints are violated considering the updated parameters. For instance, if a human agent introduces a preference not to perform a future task that was previously assigned to him/her, the constraints in~\eqref{eq:preferences} will no longer be satisfied, requiring to compute a new optimal solution to~\eqref{eq:prob}. Second, we check the allocation optimality variation. 
Let  $\mathcal{T}^+\subset \mathcal{T}$ be the  subset of tasks which still need to be performed, and let $\hat{C}^+$ and ${C}^+$ be the cost functions associated with the tasks in $\mathcal{T}^+$ and computed using the nominal and updated parameters, respectively. We reallocate if the following condition occurs: 
\begin{equation}\label{eq:delta}
     {|\hat{C}^+ - {C}^+|}\,/\,{\hat{C}^+}>\delta_t,
\end{equation}
with $\delta_t$ a positive constant.
Finally, online reallocation is applied if new tasks to execute are requested or if the available resources change. 
Note that we assume that  tasks already started cannot be interrupted in reallocation, i.e., we impose that their allocation is preserved during replanning. 


\section{Validation results}

\subsection{Setup description}\label{sec:case-study}
 The experimental setup, shown in Figure~\ref{fig:scenario}, is composed of two 7-DOFs Kinova Jaco$^2$ manipulators ($a_{r,1}$ and $a_{r,2}$), fixed on a desk, and a human operator. Inspired by the European project CANOPIES,   
focusing on HRC in  agricultural contexts, the collaborative objective is the placement of  six bunches of grapes (numbered from $1$ to $6$) in a box, divided in two layers separated by two paper cloths (numbered as $7$ and $8$). Each object $i$ is associated with a corresponding pick-and-place task $\tau_i$, that takes the object from its initial position and places it in a given location inside the box. 
 
 \begin{psfrags}
     \psfrag{t1}[1pt][1pt][0.9][0]{{\blue $\tau_1$}}
     \psfrag{t2}[1pt][1pt][0.9][0]{\blue $\tau_2$}
     \psfrag{t3}[1pt][1pt][0.9][0]{\blue $\tau_3$}
     \psfrag{t4}[1pt][1pt][0.9][0]{\blue $\tau_4$}
     \psfrag{t5}[1pt][1pt][0.9][0]{\blue $\tau_5$}
     \psfrag{t6}[1pt][1pt][0.9][0]{\blue $\tau_6$}
     \psfrag{t7}[1pt][1pt][0.9][0]{\blue $\tau_7$}
     \psfrag{t8}[1pt][1pt][0.9][0]{\blue $\tau_8$}
     \psfrag{a1}[1pt][1pt][0.9][0]{{\color{white}$a_{r,1}$}}
     \psfrag{a2}[1pt][1pt][0.9][0]{{\color{white}$a_{r,2}$}}
     
     \psfrag{a}[1pt][1pt][0.9][0]{a)}
     \psfrag{b}[1pt][1pt][0.9][0]{b)}
     \psfrag{c}[1pt][1pt][0.9][0]{c)}
     
    \mypsfrag{8.5}{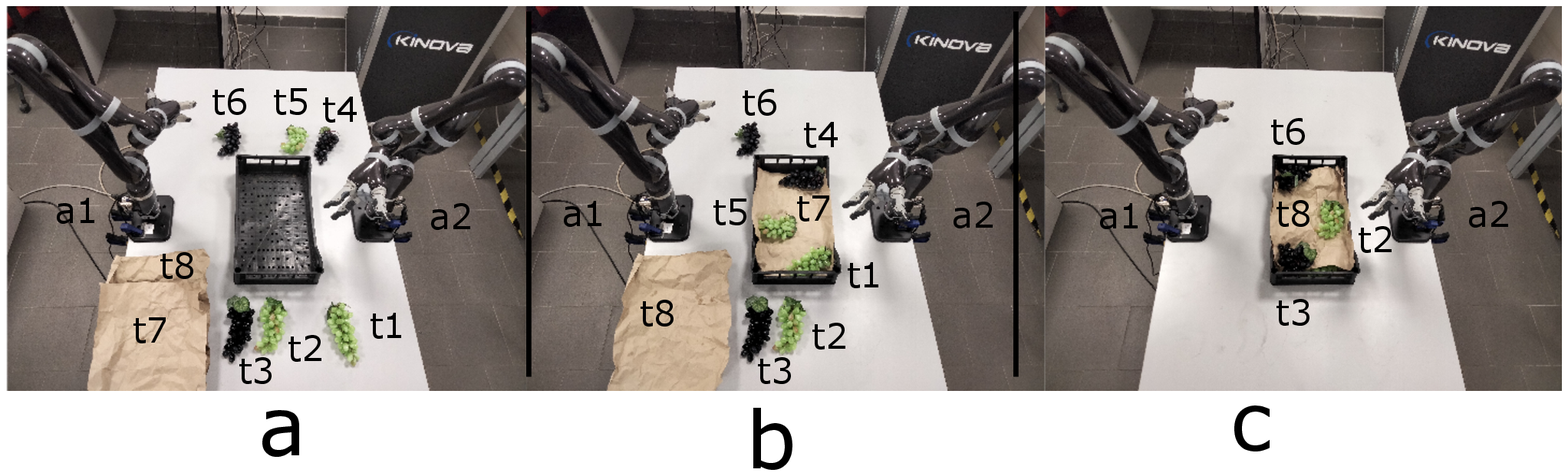}{-15pt}{Sequence of tasks to perform for composing the two layers.}{fig:scenario}
\end{psfrags}


 ROS (Robotic Operating System) middleware is used for the software architecture: one ROS node, realized in Matlab, solves the optimization problem by exploiting Gurobi solver, defines the desired trajectories of the robots and executes the checks for online reallocation, while two C++ ROS nodes realize the inverse kinematics control algorithms for the robots to follow the desired trajectories. 
 A Graphical User Interface (GUI), shown in the accompanying video, is also realized in Matlab. 
Here, the human can visualize the optimal plan, receive notifications when a task has to be started, and receive information about the current tasks executed by the robots. In addition, two buttons are displayed in the GUI that the human can press to notify the termination of a task and to express the preference not to execute a given task.

 
 
 


Figure \ref{fig:scenario}.a shows the initial positions of the objects, while Figures \ref{fig:scenario}.b  and \ref{fig:scenario}.c show the two layers to compose. The sequence of tasks to be performed is: a paper cloth needs to be placed on the box base (task $7$);  then, bunches $1$ and $4$ have to be placed close to two corners of the box, while bunch $5$ has to be positioned in the center; next, the second paper cloth is placed in the box (task 8);  finally, the last three bunches have to be released inside the box, with bunches $3$ and $6$ in the corners and $2$ in the center. 

The quality parameters $q_{i,j}$ are evaluated as a measure of the positioning accuracy in the box. 
In this perspective, the eight tasks are divided into three clusters: the first one, $\mathcal{C}_1$, includes $\tau_1$, $\tau_3$, $\tau_4$ and $\tau_6$, which require more precision as the release positions are in the corners of the box; the second one, $\mathcal{C}_2$,  is composed of $\tau_7$ and $\tau_8$, which require significant dexterity to be executed; finally, $\tau_2$ and $\tau_5$ are included in the third cluster $\mathcal{C}_3$, which gathers the tasks that do not require particularly high skills to be carried out. 
Estimated execution times for the robots are computed by using average linear velocity equal to $0.20$~m/s and average angular velocity equal to $1.7$~rad/s.
 Regarding the human agent, based on prior tests, 
they are set by considering average velocity equal to $0.20$~m/s.
In this scenario, switching costs model the time needed by the agents to go from the {\textit{place}} position of a task to the {\textit{pick}} position of the subsequent task. 

The following precedence constraints are introduced: the first paper cloth ($\tau_7$) has to be placed before the start of $\tau_1$, $\tau_4$ and $\tau_5$ (related to grape bunches), i.e., $P_{7,k}=1$ with $k\in\{1,4,5\}$; then, the second paper cloth ($\tau_8$)
has to be placed  before the start of the remaining tasks related to  grape bunches ($\tau_2$, $\tau_3$ and  $\tau_6$), i.e., $P_{8,k}=1$ with $k\in\{2,3,6\}$.
Spatial constraints are introduced between $\{\tau_4, \tau_5\}$ and $\{\tau_2, \tau_3\}$, i.e., $D_{4,5}=D_{5,4}=D_{2,3}=D_{3,2}=1$, for which initial positions are close to each other. 

The minimum quality threshold $\underline{q}$ in \eqref{eq:qual} is set to $0.8$.
Regarding robotic agents $a_{r,1}$ and $a_{r,2}$, the execution quality is set to $0.6$ for tasks of cluster $\mathcal{C}_1$, since the grape bunches have to be precisely placed in the box corners and the robots might not be able to reach this accuracy,  to $1$  for tasks in $\mathcal{C}_3$, meaning that the robots can confidently perform these tasks, and to  $0$ for tasks in $\mathcal{C}_2$, that would require too much dexterity to be performed by the robotic agents. Similarly, the execution times for tasks in $\mathcal{C}_2$ are set to $M$ for the robots. 
Concerning the human agent $a_{h,1}$, all the execution  qualities are set to $0.8$, while the  supervision ones to $1$.
The execution workloads for the all the agents are set to $1$, while the supervision workload for the human is set to $0.3$.
 
\subsection{Numerical validation campaign }
A validation campaign has been carried out where results with and without   the optimization of the switching cost are compared. In detail, the initial configurations of the $8$ objects in the scene are uniformly randomized in the interval \mbox{$[-1,1]$ m} for the $x,y$ components with respect to the center of the table. Then, for each  configuration, the distance of the objects from the center is gradually increased up to $10$ times from the initial one. 
The difference between the makespan, i.e., the total process time, resulting  without considering the switching cost and with optimizing it is computed for each test. When no  switching cost optimization is applied, robots and human are required to reach a default configuration close to the work area  after every task as in \cite{LIPPI_ROMAN2021}. Figure~\ref{fig:sw_cost_valid} shows the average and standard variation of the makespan gap obtained with $50$ random initial configurations. The figure makes evident that the mean difference is always positive, i.e., benefits in terms of makespan always occur when  considering the switching time, and increases with higher  distances (for the sake of completeness, the difference is always positive). This is motivated by the fact that the more the objects are scattered in the environment, the more crucial it is to optimize the time to change task, and thus to reach the next allocated object. For the sake of space, the other optimization variables are not reported since no significant  variation occurs for these.

\begin{psfrags}
    \def\scalnum{0.6}
    \def\scal{0.8}
    \psfrag{-100}[cc][][\scalnum]{ $-100$}
\psfrag{-50}[cc][][\scalnum]{ $-50$}
\psfrag{-40}[cc][][\scalnum]{ $-40$}
\psfrag{-30}[cc][][\scalnum]{ $-30$}
\psfrag{-20}[cc][][\scalnum]{ $-20$}
\psfrag{-4}[cc][][\scalnum]{ $-4$}
\psfrag{-2}[cc][][\scalnum]{ $-2$}
\psfrag{-0.4}[cc][][\scalnum]{ $-0.4$}
\psfrag{-0.3}[cc][][\scalnum]{ $-0.3$}
\psfrag{-0.2}[cc][][\scalnum]{ $-0.2$}
\psfrag{-0.1}[cc][][\scalnum]{ $-0.1$}
\psfrag{0}[cc][][\scalnum]{ $0$}
\psfrag{0.002}[cc][][\scalnum]{$0.002$}
\psfrag{0.004}[cc][][\scalnum]{$0.004$}
\psfrag{0.006}[cc][][\scalnum]{$0.006$}
\psfrag{0.008}[cc][][\scalnum]{$0.008$}
\psfrag{0.005}[cc][][\scalnum]{ $0.005$}
\psfrag{0.01}[cc][][\scalnum]{ $0.01$}
\psfrag{0.012}[cc][][\scalnum]{ $0.012$}
\psfrag{0.015}[cc][][\scalnum]{ $0.015$}
\psfrag{0.02}[cc][][\scalnum]{$0.02$}	
\psfrag{0.025}[cc][][\scalnum]{$0.025$}	
\psfrag{0.04}[cc][][\scalnum]{$0.04$}	
\psfrag{0.06}[cc][][\scalnum]{$0.06$}	
\psfrag{0.08}[cc][][\scalnum]{$0.08$}
\psfrag{0.05}[cc][][\scalnum]{ $0.05$}
\psfrag{0.1}[cc][][\scalnum]{ $0.1$}
\psfrag{0.2}[cc][][\scalnum]{ $0.2$}
\psfrag{0.4}[cc][][\scalnum]{ $0.4$}
\psfrag{0.6}[cc][][\scalnum]{ $0.6$}
\psfrag{0.8}[cc][][\scalnum]{ $0.8$}
\psfrag{0.82}[cc][][\scalnum]{ $0.82$}
\psfrag{0.83}[cc][][\scalnum]{ $0.83$}
\psfrag{0.84}[cc][][\scalnum]{ $0.84$}
\psfrag{0.85}[cc][][\scalnum]{ $0.85$}
\psfrag{0.86}[cc][][\scalnum]{ $0.86$}
\psfrag{0.5}[cc][][\scalnum]{ $0.5$}	\psfrag{1}[cc][][\scalnum]{ $1$}
\psfrag{1.5}[cc][][\scalnum]{ $1.5$}
\psfrag{2}[cc][][\scalnum]{ $2$}
\psfrag{2.5}[cc][][\scalnum]{ $2.5$}
\psfrag{3}[cc][][\scalnum]{ $3$}
\psfrag{4}[cc][][\scalnum]{ $4$}
\psfrag{4.5}[cc][][\scalnum]{ $4.5$}
\psfrag{5}[cc][][\scalnum]{ $5$}
\psfrag{5.2}[cc][][\scalnum]{ $5.2$}
\psfrag{5.4}[cc][][\scalnum]{ $5.4$}
\psfrag{5.6}[cc][][\scalnum]{ $5.6$}
\psfrag{5.8}[cc][][\scalnum]{ $5.8$}
\psfrag{6}[cc][][\scalnum]{ $6$}
\psfrag{7}[cc][][\scalnum]{ $7$}
\psfrag{8}[cc][][\scalnum]{ $8$}
\psfrag{9}[cc][][\scalnum]{ $9$}
\psfrag{10}[cc][][\scalnum]{ $10$}
\psfrag{11}[cc][][\scalnum]{ $11$}
\psfrag{12}[cc][][\scalnum]{ $12$}
\psfrag{13}[cc][][\scalnum]{ $13$}
\psfrag{14}[cc][][\scalnum]{ $14$}
\psfrag{15}[cc][][\scalnum]{ $15$}
\psfrag{16}[cc][][\scalnum]{ $16$}
\psfrag{18}[cc][][\scalnum]{ $18$}
\psfrag{20}[cc][][\scalnum]{ $20$}
\psfrag{25}[cc][][\scalnum]{ $25$}
\psfrag{26}[cc][][\scalnum]{ $26$}
\psfrag{28}[cc][][\scalnum]{ $28$}
\psfrag{30}[cc][][\scalnum]{ $30$}
\psfrag{32}[cc][][\scalnum]{ $32$}
\psfrag{34}[cc][][\scalnum]{ $34$}
\psfrag{35}[cc][][\scalnum]{ $35$}
\psfrag{36}[cc][][\scalnum]{ $36$}
\psfrag{38}[cc][][\scalnum]{ $38$}
\psfrag{40}[cc][][\scalnum]{ $40$}
\psfrag{50}[cc][][\scalnum]{ $50$}
\psfrag{60}[cc][][\scalnum]{$60$}
\psfrag{70}[cc][][\scalnum]{$70$}
\psfrag{80}[cc][][\scalnum]{$80$}
\psfrag{90}[cc][][\scalnum]{$90$}
\psfrag{100}[cc][][\scalnum]{$100$}
\psfrag{120}[cc][][\scalnum]{$120$}
\psfrag{140}[cc][][\scalnum]{$140$}
\psfrag{160}[cc][][\scalnum]{$160$}
\psfrag{150}[cc][][\scalnum]{$150$}
\psfrag{180}[cc][][\scalnum]{$180$}
\psfrag{200}[cc][][\scalnum]{$200$}
\psfrag{250}[cc][][\scalnum]{$250$}
\psfrag{300}[cc][][\scalnum]{$300$}
\psfrag{350}[cc][][\scalnum]{$350$}
\psfrag{400}[cc][][\scalnum]{$400$}
\psfrag{500}[cc][][\scalnum]{$500$}
\psfrag{600}[cc][][\scalnum]{$600$}
\psfrag{700}[cc][][\scalnum]{$700$}
\psfrag{800}[cc][][\scalnum]{$800$}
\psfrag{900}[cc][][\scalnum]{$900$}
\psfrag{1000}[cc][][\scalnum]{$1000$}
\psfrag{1200}[cc][][\scalnum]{$1200$}
\psfrag{1400}[cc][][\scalnum]{$1400$}
\psfrag{1600}[cc][][\scalnum]{$1600$}
\psfrag{1800}[cc][][\scalnum]{$1800$}
\psfrag{2000}[cc][][\scalnum]{$2000$}
\psfrag{-0.5}[cc][][\scalnum]{ $\!\!\!-0.5$}
\psfrag{-1}[cc][][\scalnum]{ $\!\!\!-1$}
\psfrag{-1.5}[cc][][\scalnum]{ $\!\!\!-1.5$}
\psfrag{-2.6}[cc][][\scalnum]{ $\!\!\!-2.6$}
\psfrag{-2.8}[cc][][\scalnum]{ $\!\!\!-2.8$}
\psfrag{-3}[cc][][\scalnum]{ $\!\!\!-3$}
\psfrag{-3.2}[cc][][\scalnum]{ $\!\!\!-3.2$}
\psfrag{-5}[cc][][\scalnum]{ $\!\!\!-5$}
\psfrag{-10}[cc][][\scalnum]{ $\!\!\!-10$}

     \psfrag{x}[1pt][1pt][\scal][0]{Distance increase factor}
     \psfrag{y}[1pt][1pt][\scal][0]{Makespan gap [s]{\vspace{1cm}}}
    \mypsfrag{7.8}{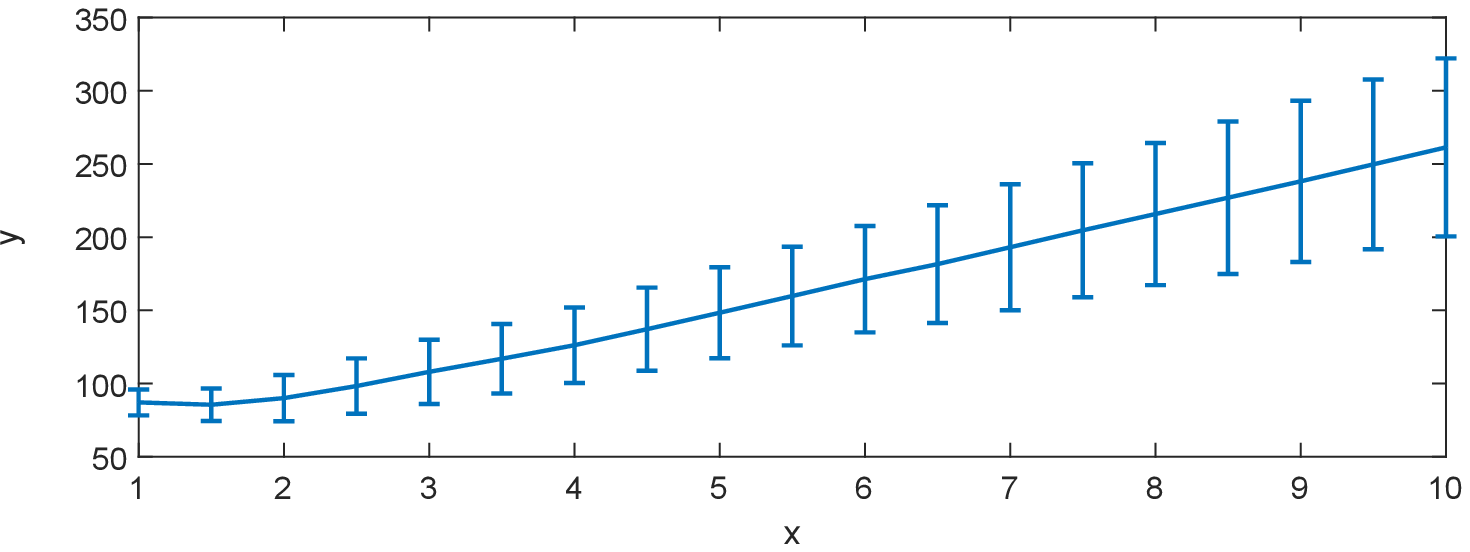}{-10pt}{Makespan gap obtained without and with handling the switching cost. 
    Average and standard deviation values are reported. }{fig:sw_cost_valid}
\end{psfrags}

\subsection{Experimental results}
 
  We now present the experimental results of the proposed framework. 
  A video reporting the complete experiments is provided as supplementary material. 
 
\subsubsection{Optimal plan and switching cost comparison}
First, we show in Figure~\ref{fig:switching}-top the optimal plan that is obtained by solving problem~\eqref{eq:prob} for the considered setup. 
Tasks assigned to each agent ($a_{r,1}, a_{r,2}, a_{h,1}$) are depicted as segments representing the task duration, where solid segments indicate task executions, and dashed ones denote task supervisions. The figure shows that the computed plan meets all the constraints, and assigns, among others, tasks $\tau_7$ and $\tau_8$ for paper cloths to the human since they cannot be executed by the robots. The solution reaches total cost equal to $0.47$, with makespan equal to $84.57$~s. To execute the plan, as shown in the accompanying video, this is first displayed on the GUI. When the human presses a {\textit{START}} button, the process begins. At this point, he executes $\tau_7$ and presses the {\textit{FINISHED}} button to notify the task termination. Then, he supervises $a_{r,2}$ during the execution of $\tau_1$, while $a_{r,1}$ executes $\tau_5$. Once these tasks are completed, the robot $a_{r,2}$ starts the execution of $\tau_4$ under human supervision. Following the plan, after the execution of $\tau_8$ by the human, agent $a_{r,1}$ starts $\tau_3$ under supervision, while finally $a_{r,1}$ executes $\tau_2$ and $a_{h,1}$ executes $\tau_6$.

For the sake of comparison, and similarly to the above, the optimization problem is also solved without taking into account the switching cost. The resulting plan is shown in Figure \ref{fig:switching}-bottom and its execution is reported in the video. The figure shows that, again, all the requirements are met. However, in this case the solution reaches a much higher total cost, namely  $0.70$, which is mainly due to a significantly higher makespan, equal to  $111.23$~s. This reconfirms the effectiveness of the proposed formulation.

\begin{psfrags}
    \def\scalnum{0.6}
    \def\scal{0.8}
    
     \psfrag{1}[1pt][1pt][\scal][0]{$a_{r,1}$}
     \psfrag{2}[1pt][1pt][\scal][0]{$a_{r,2}$}
     \psfrag{3}[1pt][1pt][\scal][0]{$a_{h,1}$}
     \psfrag{t}[1pt][1pt][\scal][0]{$t$[s]}
    \psfrag{a}[cc][][\scalnum]{ $\A$}

    \psfrag{l}[1pt][1pt][1][0]{}
     \psfrag{t1}[1pt][1pt][1][0]{$\tau_1$}
     \psfrag{t2}[1pt][1pt][1][0]{$\tau_2$}
     \psfrag{t3}[1pt][1pt][1][0]{$\tau_3$}
     \psfrag{t4}[1pt][1pt][1][0]{$\tau_4$}
     \psfrag{t5}[1pt][1pt][1][0]{$\tau_5$}
     \psfrag{t6}[1pt][1pt][1][0]{$\tau_6$}
     \psfrag{t7}[1pt][1pt][1][0]{$\tau_7$}
     \psfrag{t8}[1pt][1pt][1][0]{$\tau_8$}

     \psfrag{Without switching cost}[1pt][1pt][0.8][0]{\small{Without switching cost}}
      \psfrag{With switching cost}[1pt][1pt][0.8][0]{\small{With switching cost}}

     \begin{figure}[!htbp]\label{fig:scenari}
        \begin{center}
        \def\arraystretch{0.1}
            \begin{tabular}[h]{c}
                         {\leavevmode{\includegraphics[width=7.8truecm]{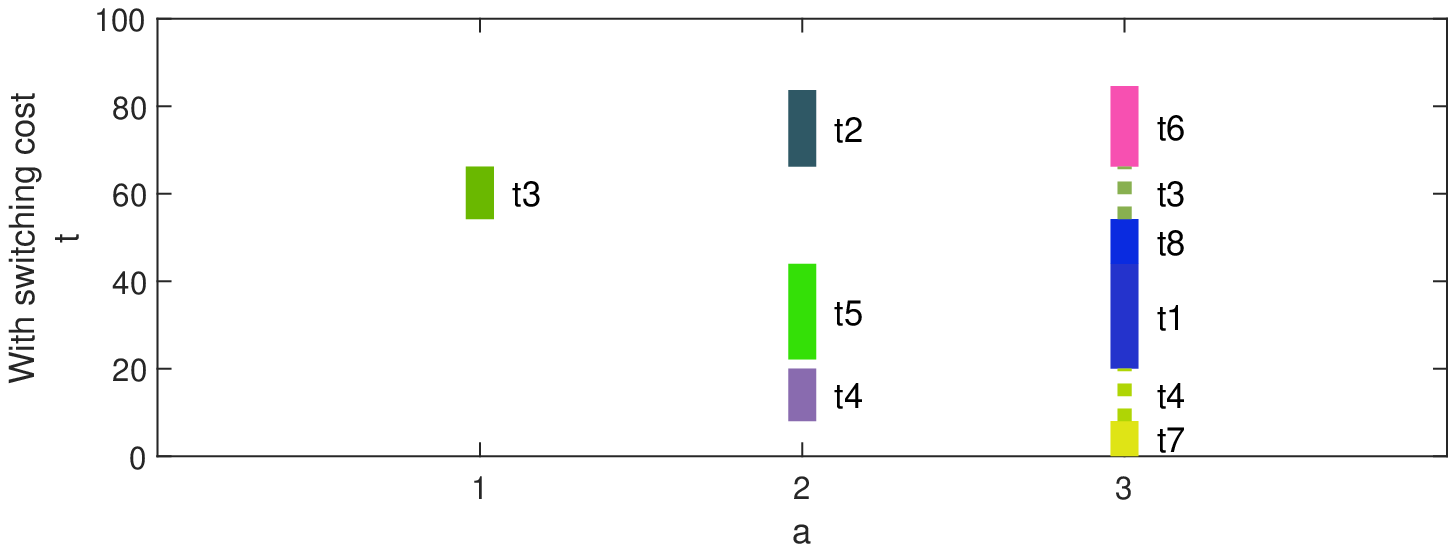}}}
                        \\  {\leavevmode{\includegraphics[width=7.8truecm]{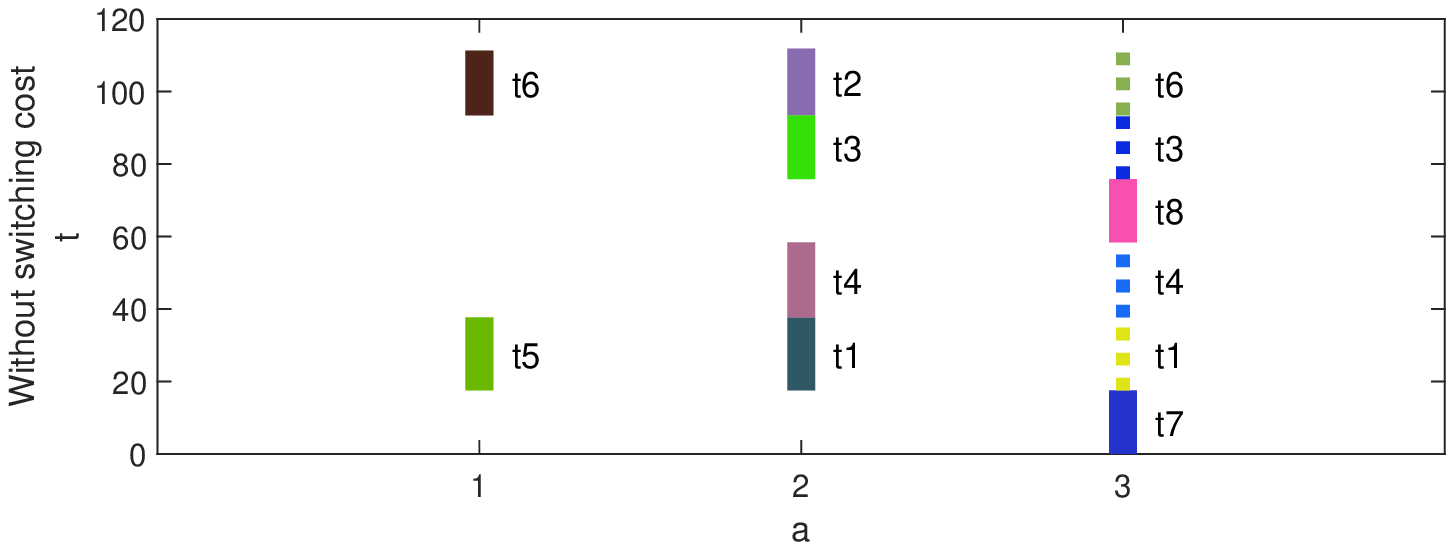}}}
             
         \end{tabular}
          \vspace{-6pt}
          \caption{Task allocation obtained by taking into account the switching costs (top) and without considering them (bottom). }
          \label{fig:switching}
        \end{center}
\end{figure}

\end{psfrags}


\subsubsection{Reallocation} 
In this case study we consider that the human expresses online the preference not to perform an assigned task. In detail, starting from the plan in Figure \ref{fig:switching}-top and after the execution of task $\tau_7$, the human asks (using the GUI) not to execute his next task, i.e., $\tau_1$, to reduce his load. This preference is added to the set $\mathcal{H}_p$ and the reallocation strategy in Sec.~\ref{sec:reall-b}   is followed. Since the preference leads the constraint~\eqref{eq:preferences} to no longer be satisfied, a new optimal plan is computed which redistributes the tasks as depicted in Figure~\ref{fig:scenari}. This plan is then shown to the human  and the  robot trajectories  are updated accordingly. It 
is worth noticing that the tasks that were already active at the time of the reallocation request ($\tau_4$ for agent $a_{r,1}$, supervised by agent $a_{h,1}$) are not affected by the reallocation. 
The updated plan is finally completed as shown in the accompanying video.




\begin{psfrags}
    \def\scalnum{0.6}
    \def\scal{0.8}
    
     \psfrag{1}[1pt][1pt][\scal][0]{$a_{r,1}$}
     \psfrag{2}[1pt][1pt][\scal][0]{$a_{r,2}$}
     \psfrag{3}[1pt][1pt][\scal][0]{$a_{h,1}$}
     \psfrag{t}[1pt][1pt][\scal][0]{$t$[s]}
    \psfrag{a}[cc][][\scalnum]{ $\A$}

    \psfrag{l}[1pt][1pt][1][0]{}
     \psfrag{t1}[1pt][1pt][1][0]{$\tau_1$}
     \psfrag{t2}[1pt][1pt][1][0]{$\tau_2$}
     \psfrag{t3}[1pt][1pt][1][0]{$\tau_3$}
     \psfrag{t4}[1pt][1pt][1][0]{$\tau_4$}
     \psfrag{t5}[1pt][1pt][1][0]{$\tau_5$}
     \psfrag{t6}[1pt][1pt][1][0]{$\tau_6$}
     \psfrag{t7}[1pt][1pt][1][0]{$\tau_7$}
     \psfrag{t8}[1pt][1pt][1][0]{$\tau_8$}
     
       \psfrag{y}[1pt][1pt][0.8][0]{Online reallocation}

     \begin{figure}[!htbp]\label{fig:scenari}
        \begin{center}
              {\leavevmode{\includegraphics[width=7.8truecm]{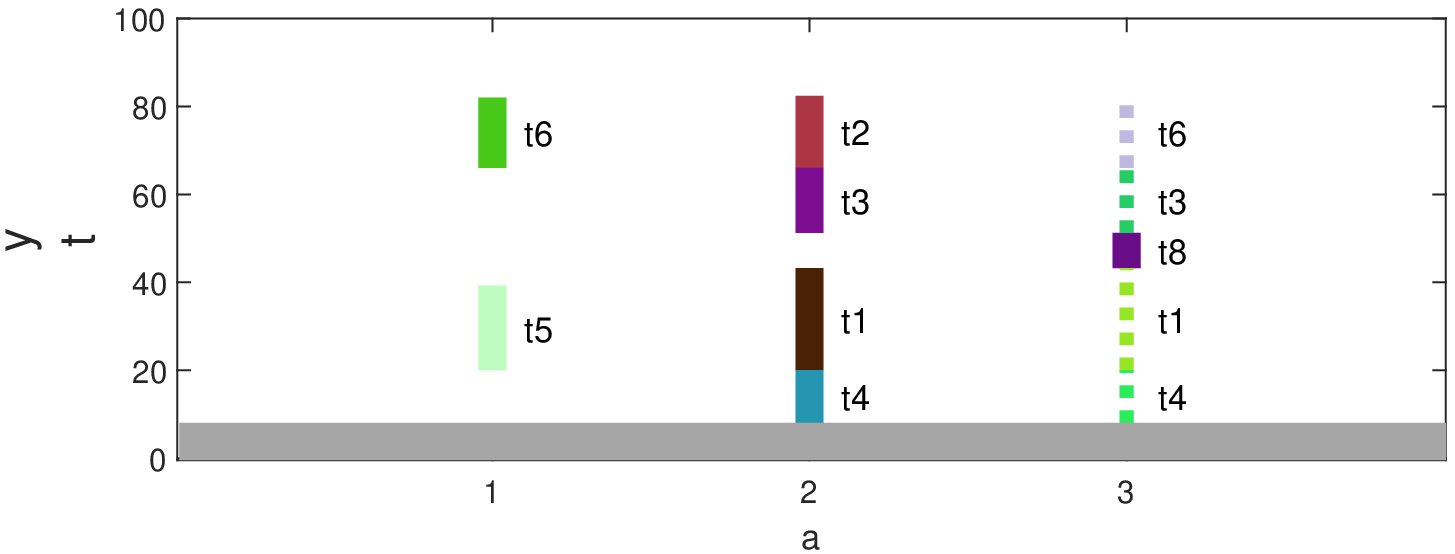}}}
          \vspace{-6pt}
          \caption{Task allocation obtained after the online reallocation. }
          \vspace{-5pt}
          \label{fig:scenari}
        \end{center}
\end{figure}

\end{psfrags}

\section{Conclusions}
In this paper,  a task allocation framework for human-robot teams was devised. The proposed solution is based on \emph{i)} an optimal offline allocation that assigns tasks among agents while taking into account their inherently different natures, and \emph{ii)} a re-planning strategy that accounts for time-varying parameters and human preferences.  
The solution was validated via experiments on a setup involving two robots and a human operator. Future works will be aimed at extending the framework with a human activity prediction module to endow the robots with proactive behaviors.

 \bibliography{biblio}
\end{document}